\documentclass{bmvc2k}

\title{Geometry Driven Progressive Warping for One-Shot Face Animation}

\addauthor{Yatao Zhong}{yazhong@microsoft.com}{1}
\addauthor{Faezeh Amjadi}{faamja@microsoft.com}{1}
\addauthor{Ilya Zharkov}{zharkov@microsoft.com}{1}

\addinstitution{
 Applied Sciences, Microsoft\\
 Redmond, WA, USA
}

\runninghead{Zhong, Amjadi and Zharkov}{Geometry Driven Progressive Warping}

\usepackage{amsthm}
\usepackage{paralist}
\usepackage{booktabs}
\usepackage{multirow}
\usepackage{array}
\usepackage{wrapfig}
\usepackage[10pt]{moresize}

\begin{document}

\maketitle

\begin{abstract}
Face animation aims at creating photo-realistic portrait videos with animated poses and expressions. A common practice is to generate displacement fields that are used to warp pixels and features from source to target. However, prior attempts often produce sub-optimal displacements. In this work, we present a geometry driven model and propose two geometric patterns as guidance: 3D face rendered displacement maps and posed neural codes. The model can optionally use one of the patterns as guidance for displacement estimation. To model displacements at locations not covered by the face model (e.g., hair), we resort to source image features for contextual information and propose a progressive warping module that alternates between feature warping and displacement estimation at increasing resolutions. We show that the proposed model can synthesize portrait videos with high fidelity and achieve the new state-of-the-art results on the VoxCeleb1 and VoxCeleb2 datasets for both cross identity and same identity reconstruction.
\end{abstract}


\section{Introduction} \label{sec:intro}
\begin{figure}[t]
\vspace{-0.02\hsize}
  \scriptsize
  \centering
    \subfigure[]
  {
    \label{fig:pull}
    \includegraphics[width=0.32\textwidth]{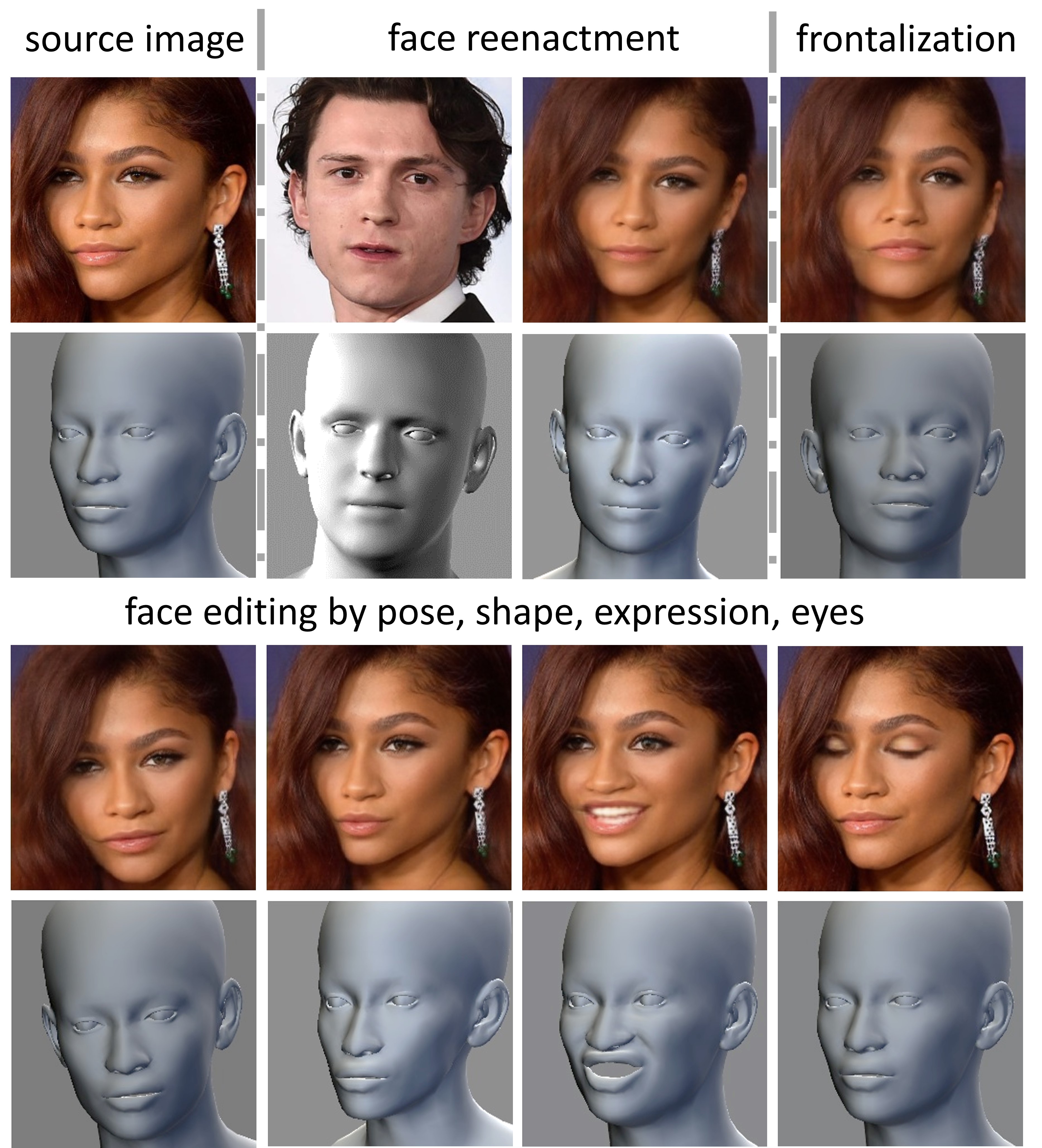}
  }
  \subfigure[]
  {
    \label{fig:model_struct}
    \includegraphics[width=0.63\textwidth]{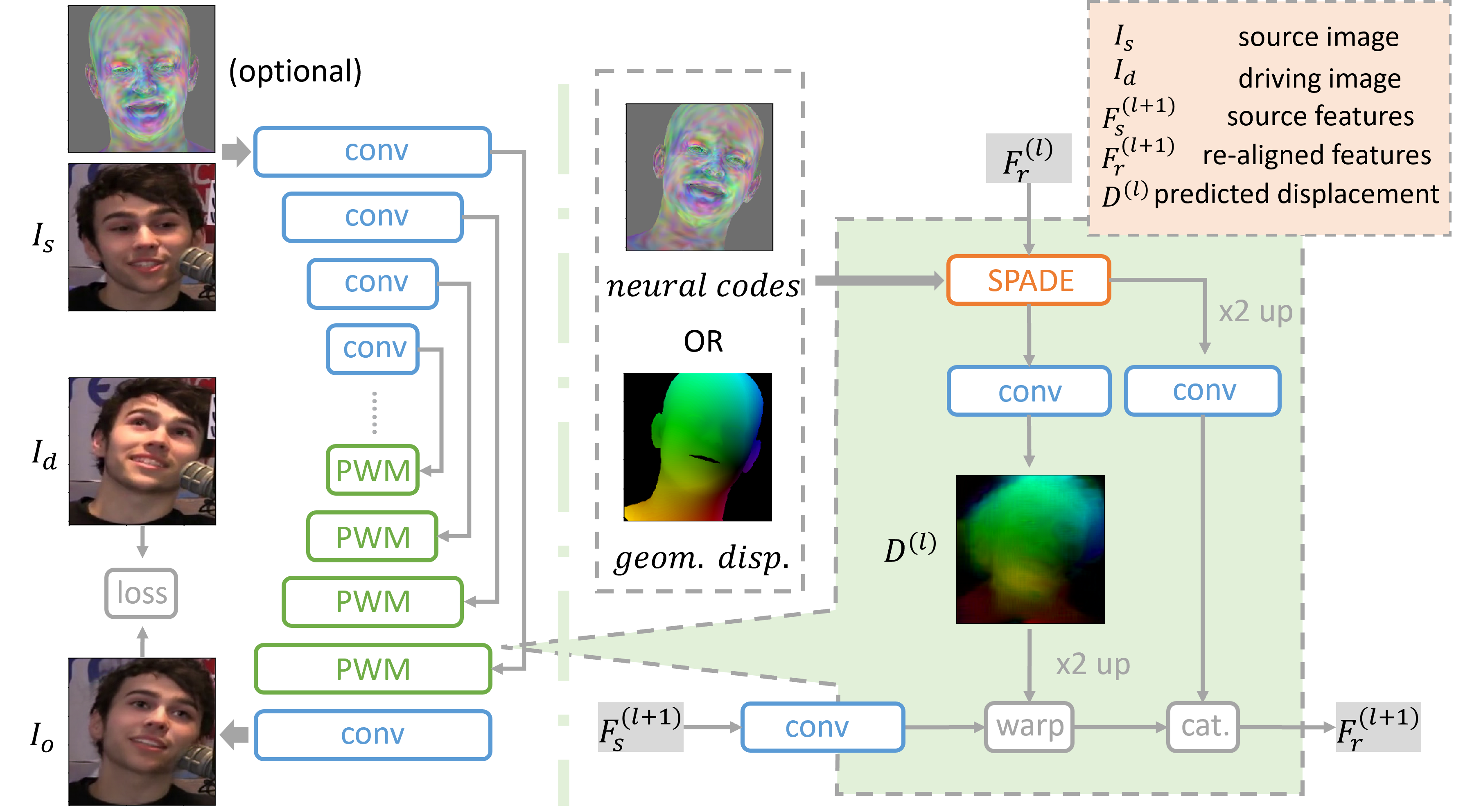}
  }
  \caption{(a) Our model can be applied to several scenarios: face reenactment, face frontalization and fine-grained face editing by pose, shape, expression and even eyes. (b) An overview of the model. The left part shows the whole architecture. The right part shows the proposed progressive warping module (PWM) at pyramid level $l$ and how we estimate the full displacement from the guiding geometric patterns.}
\vspace{-0.05\hsize}
\end{figure}

Face animation refers to the task of creating photo-realistic portrait videos with animated facial motions. The task starts with a source portrait image and a sequence of driving face poses and expressions. For each frame in the output video, a face animation model generates a new portrait image with the same pose and expression as the driving face while still preserving the source identity and appearance (see examples in Fig. \ref{fig:pull}). In the early days of research, solutions often rely on purely graphics and geometry \cite{dale2011video,garrido2014automatic,thies2016face2face,thies2018headon}. They utilize face geometries such as facial landmarks and reconstructed 3D face models to guide face deformation. However, image synthesis in those early works is non-learning based and often resorts to handcrafted image blending heuristics, hence fails to produce photo-realistic output.

Recently with the rapid advance in image synthesis using deep neural networks, many solutions switch to a learning-based paradigm where they directly operate in the feature space of source images and warp features to achieve facial motion transfer and identity preservation. However, prior works \cite{doukas2021headgan,ren2021pirenderer,zhang2021flow} attempt to predict the warping displacements directly from features in source image coordinates whereas the displacements are supposed to be in the target image coordinates. This misalignment issue poses a challenge to them to generate correct displacements. 

With those considerations in mind, we ask how we can improve the quality of displacement fields for face animation? We present a geometry driven model and propose two geometric patterns as guidance: 3D face rendered displacement maps and posed neural codes. The model can optionally use one of the patterns as guidance for displacement estimation. To render the displacement maps, we first calculate the displacement vectors between corresponding vertices from a pair of source and driving meshes and then rasterize them to a $2$-channel image. To create posed neural codes, we attach a latent embedding vector to each vertex of the 3D face and rasterize the embedded mesh to a $d$-channel image, where $d$ is the embedding dimension. For choice of the 3D face model, we use FLAME \cite{li2017learning}, which is a face counterpart of the blend-skinned body model SMPL \cite{loper2015smpl}. Unlike other 3DMMs such as the FaceWarehouse \cite{cao2013facewarehouse} and the Basel Face Model \cite{paysan20093d}, FLAME also models articulated jaw, neck and rotating eyeballs in addition to the disentangled shape, pose and expression parameters, which makes it a more natural representation for face animation. To overcome the spatial misalignment issue present in other 3DMM based models \cite{doukas2021headgan,ren2021pirenderer,zhang2021flow}, we warp source image features with the predicted displacement field at each pyramid scale and use the warped features to estimate the displacement field at next pyramid scale. We repeat this paradigm and progressively warp features and estimate displacements at increasing resolutions. 

To summarize, our work has the following contributions:
\begin{inparaenum}[1)]
\item We propose a geometry driven model whose displacement generation is guided by either displacement maps or posed neural codes, both rendered with the FLAME face topology.
\item We design a progressive warping module which alternates between displacement estimation and feature warping for robust facial motion transfer.
\item We demonstrate that our method sets the new-state-of-art through extensive experiments and provide our insight into the properties of the proposed guiding geometric patterns by analyzing their individual impact on the model performance.
\end{inparaenum}

\section{Related Work}
\textbf{3D Face Based Warping Models.} 
In early research of face animation, solutions often rely on purely 3D face models. For instance, \cite{dale2011video,thies2016face2face,thies2018headon} start with 3D face reconstruction and use parametric face models to control the face movement. However, image synthesis in those early works is non-learning based and often resorts to some heuristic blending techniques, hence fails to produce photo-realistic output. Later come the learning-based solutions. \cite{kim2018deep,yi2020audio} first render synthetic face images with predicted texture and lighting, which are then used as input to deep networks to create images with better realism. PIRender \cite{ren2021pirenderer} encodes the 3DMM pose parameters into an embedding vector and uses that as modulation in AdaIN layers \cite{huang2017arbitrary} to estimate the displacement field for face deformation. HeadGAN \cite{doukas2021headgan} shares a similar idea but uses a 3D face image as the source of modulation in SPADE layers \cite{park2019semantic} for displacement generation. \cite{zhang2021flow} creates an approximate flow with the 3D face model for initial warping and the warped image is refined by another generator to create a more realistic image. 

\textbf{Keypoint Based Warping Models.}
Warp-Guided GANs \cite{geng2018warp} track the facial landmarks and use them to create a global displacement map for image warping. Subsequently the warped image is fed to a face refinement network to generate final image. Facial landmarks are also used in MarioNETte \cite{ha2020marionette} to create a global displacement map, but here warping operates in the feature space instead of image pixels. Another line of work under this category includes an ad hoc keypoint detector in the model, which is trained without direct supervision \cite{siarohin2019animating,siarohin2019first,wang2021one}. The keypoints are implicitly learned by the model and may not convey any interpretable meaning to humans. The displacement between each pair of keypoints controls a local motion transformation. A dense motion network is used to predict the weighting coefficient of each keypoint displacement in order to create a global warping field. MonkeyNet \cite{siarohin2019animating} is one of the first works that go in the direction of latent keypoint representation. Follow-up works such as FOMM \cite{siarohin2019first} and 3D-FOMM \cite{wang2021one} improve MonkeyNet \cite{siarohin2019animating} with first-order motion modeling and extend it from 2D to 3D respectively.

\textbf{Pose-to-Face Mapping Models.}
Pose-to-Face refers to a class of models that transform input directly from pose encoding to face images without considering the geometry such as facial landmarks or 3D face models. They share the same spirit of pix2pix \cite{isola2017image} and vid2vid \cite{wang2018video, wang2019few}.
X2Face \cite{wiles2018x2face} is one of the first works in this direction. They have an embedding network that encodes the source identity and a driving network that uses the driving pose and embedded source face to synthesize the new face image. Bi-layer model \cite{zakharov2020fast} shares the similar idea of embedded face, but they have a two-stage implementation for coarse-to-fine refinement. \cite{zakharov2019few} encodes an image drawn with landmarks of driving face and uses the source image features injected in AdaIN \cite{huang2017arbitrary} to generate output. Similar to \cite{zakharov2019few, zakharov2020fast}, LSR \cite{meshry2021learned} is also driven by a landmark encoded image, but LSR uses a second network trained for semantic segmentation to guide the decoder to generate better quality images. Head2Head \cite{koujan2020head2head} proposes a person-specific model that goes directly from driving 3D face to output image without any feature extraction from source identity. Therefore, their model does not generalize. 


\vspace{-0.02\hsize}
\section{Proposed Method}
We start by fitting the FLAME face model \cite{li2017learning} to the input images (Sec. \ref{sec:3d_face_repr}). To embed FLAME in our model and utilize its geometric information to guide facial motion transfer, we use one of the geometric patterns: the FLAME mesh rendered displacement map or the posed neural codes. Either acts as guidance to estimate the full displacement field used for feature warping (Sec. \ref{sec:learn_disp}). Afterwards we apply a novel progressive warping module and alternate between feature warping and displacement estimation to generate output images. (Sec. \ref{sec:pwm}). 

\subsection{3D Face Representation} \label{sec:3d_face_repr}
To obtain full control over the animated facial motions, we need a latent face descriptor that is compact yet expressive. In this work, we employ FLAME \cite{li2017learning} as the underlying face representation due to its capability of modeling articulated joints (jaw, neck and eyeballs) and disentangled parameterization of shape $\beta$, pose $\theta$ and expression $\psi$. FLAME deforms face geometry by vertex based linear blend skinning (LBS) with corrective blendshapes. We refer readers to \cite{li2017learning} for detailed explanation of the FLAME formulation.

Given a source portrait image $I_s$ that we want to animate, we use DECA \cite{feng2021learning}, an off-the-shelf 3D face reconstruction model, to fit the FLAME face geometry. The output of the fitting pipeline is the shape $\beta_s$, pose $\theta_s$ and expression $\psi_s$ for the source image $I_s$, with which we can reconstruct a FLAME mesh $M_s$. In order to animate the source face with customized facial movement, we can change the values of $\beta_s$, $\theta_s$ and $\psi_s$ to obtain the target shape $\beta_d$, pose $\theta_d$ and expression $\psi_d$. By substituting the updated parameters into the FLAME model, we now have the target 3D face mesh $M_d$ that will drive the movement of source face. In the case of face reenactment where we want to transfer the facial motion from a driving portrait image $I_d$ to the source image $I_s$, we do an additional FLAME fitting to $I_d$ to generate the driving 3D face mesh $M_d$. 

\subsection{Geometric Patterns as Guidance} \label{sec:learn_disp}

Our model is driven by facial geometry. We propose two geometric patterns that can guide the displacement learning process. One is a $2$-channel image rendered from the displacement vectors between a pair of source and target face meshes, which we call the \textit{geometric displacement field}. The other is the \textit{posed neural codes}, a $d$-channel image rendered from a face mesh whose vertices are embedded with a set of $d$ dimensional latent vectors. We now describe the formulation of each of the geometric patterns. 

\textbf{Geometric displacement field}. Given the source and driving face meshes $M_s$ and $M_d$, we calculate 2D displacement vectors $V_{d \rightarrow s} = \mathcal{P}(M_s) - \mathcal{P}(M_d)$, where $\mathcal{P}(\cdot)$ projects a mesh to image space. For each vertex $i$, we take $V_{d \rightarrow s, i}$ as its vertex attribute. Since the mesh $M_d$ is already triangulated, we can follow the conventional graphics pipeline for rasterization. Note that we render the displacement vectors using the topology of driving mesh $M_d$ because our goal is to warp features from the source image coordinates and place them in the target image coordinates. We also do back-face culling to ignore any triangles that are invisible from the viewing direction.

\textbf{Posed neural codes}. Geometric displacement field can assist with the full displacement estimation but it mostly transfers the relative motion from source to target and does not explicitly incorporate facial semantics. To enforce the full displacements conditioned on facial semantics, we propose the posed neural codes.

\begin{wrapfigure}{r}{0.5\linewidth}
\vspace{-0.02\hsize}
\scriptsize
\includegraphics[width=\linewidth]{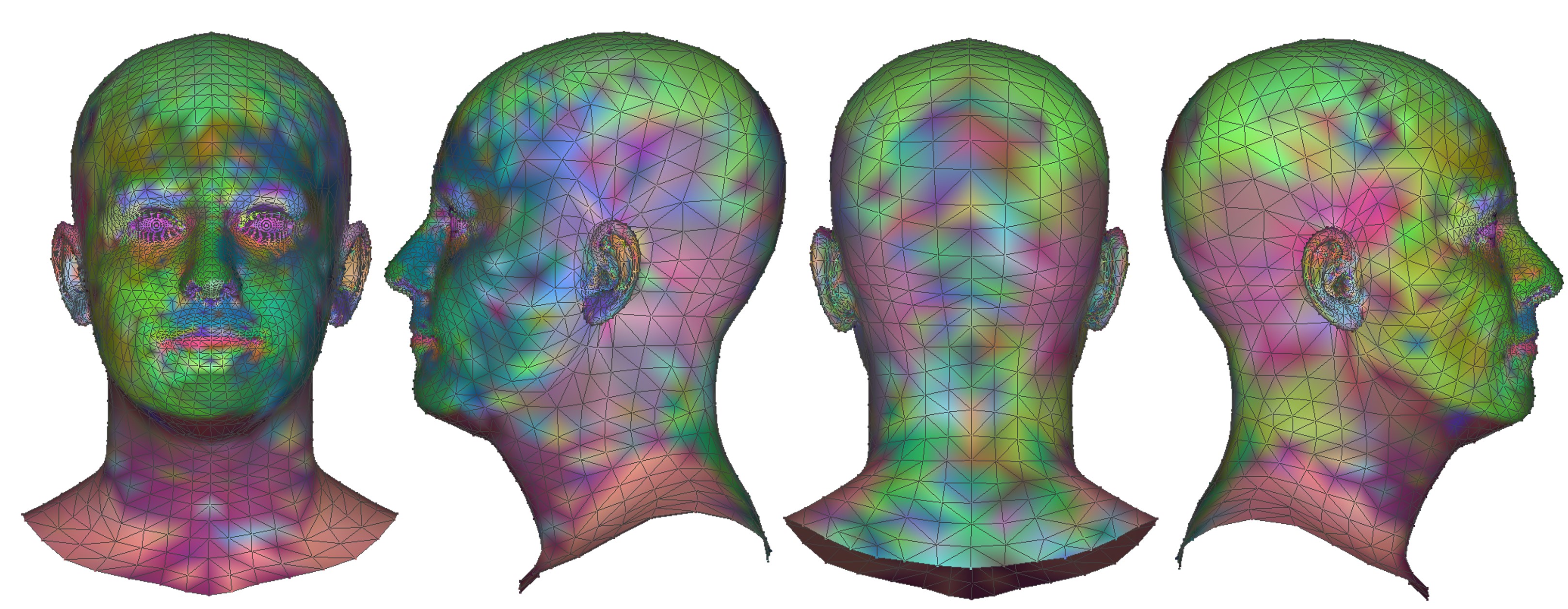}
\caption{Visualized latent codes on the FLAME manifold. Color encoding is obtained from t-SNE. Similar colors indicate closer distance in the embedding space.}
\label{fig:latent_codes_viz}
\vspace{-0.02\hsize}
\end{wrapfigure}

For a FLAME model with $N$ vertices, we define a set of $d$ dimensional latent vectors $E=\{ e_1, e_2, e_3, \cdots, e_N \}$ with $e_i \in R^d$. We attach each vector $e_i$ to vertex $i$ and obtain a mesh embedded in $R^{N \times d}$. Similar to geometric displacement field, we rasterize a $d$-channel image from the embedded mesh. Because the mesh topology is predefined and fixed, each vertex comes with semantic meaningfulness, hence the assigned latent vector too. A visualization of the learned latent codes is available in Fig. \ref{fig:latent_codes_viz}. As can be seen, the front and back both have symmetric coloring. The left and right possess similar color distribution patterns. They suggest the facial semantics being well captured by the latent codes. 

As opposed to geometric displacement field, posed neural codes does not encode relative motion directly.  To recover this information, we render the latent codes with the source and target meshes respectively and provide both as input to the model. The source rendered latent codes are concatenated with the source image as input to the encoder. The target rendered latent codes are used as guidance in the decoder (Fig. \ref{fig:model_struct}). The rendering process is fully differentiable and the latent codes are learned along with the neural network parameters.

\textbf{Learning the full displacement field}. Although both geometric patterns encode accurate facial geometry and motion transfer information, they do not model any surrounding pixels outside the face or head region (e.g. hair). To address this issue, we propose to use the geometric patterns as guidance to estimate the full displacement field. Imagine that we already have the spatially re-aligned features from a preceding warping module (which we will touch later in Sec. \ref{sec:pwm}). Since the input features have been warped to align with the driving pose in target image space, we adopt them for contextual information to fill in any missing pieces that are not modeled by the 3D face. To accomplish this, we use either the rasterized geometric displacement field or the posed neural codes as the modulation input to a SPADE block \cite{park2019semantic} to guide the estimation of full displacement field (Fig. \ref{fig:model_struct}). 

\subsection{Progressive Warping Module} \label{sec:pwm}

The goal of estimating displacements is to warp the source features and rearrange them in the target image space. However, many existing works often fail to generate displacement vectors that point to the correct source locations because of the spatial misalignment issue discussed in Sec. \ref{sec:intro}.  We propose a plug-in feature re-sampling block that can warp the source features to create new ones that are spatially aligned with the driving pose in target image coordinates. The re-sampling block is repeatedly applied at increasing pyramid scales to finally create a high-resolution displacement field. We term it the \textit{progressive warping module}, or PWM for short.

Fig. \ref{fig:model_struct} shows an overview of the model architecture. Unlike existing methods that employ a number of dedicated deep networks for feature extraction, displacement estimation, image generation and image refinement and operate in a multi-stage pipeline, our model implements a simple encoder-decoder structure. The decoder part adopts a sequence of progressive warping modules for iterative feature warping and re-alignment, which assists full displacement estimation at increasing resolutions. 

Let $L$ be the number of feature pyramid levels and $l$ the index of the $l$-th level. The PWM starts from the lowest resolution at level $l=1$ in the decoder and gradually moves to the highest resolution at $l=L$. As shown in Fig. \ref{fig:model_struct}, at each pyramid level $l$, the guidance map (geometric displacement field or posed neural codes) is resized to match the resolution at level $l$ and injected into a SPADE block \cite{park2019semantic} to modulate the re-aligned features $F_r^{(l)}$ from a preceding PWM. The SPADE output is followed by a convolutional layer to predict the full displacement field $D^{(l)}$, which is up-sampled and used to warp the shortcut features $F_s^{(l+1)}$ from encoder. The warped features are concatenated with the other route of the SPADE output to create a set of new spatially re-aligned features $F_r^{(l+1)}$, which will be used by the PWM at next level. A final convolutional layer is added on top of the last PWM to synthesize the output image $I_o$.

A sequence of $L$ PWMs are employed so that we are able to generate high resolution displacement fields and features to animate portrait images. A special case is the initial PWM at the lowest resolution, where no re-aligned features from a preceding PWM is available. We go around this by skipping the SPADE and displacement estimation components. If the guidance map is the geometric displacement field, we directly adopt it (resized to match the corresponding resolution) to warp the encoder features $F_s^{(1)}$, which are concatenated with the original $F_s^{(1)}$ to produce $F_r^{(1)}$. If the guidance map is posed neural codes, we use $F_s^{(1)}$ to produce $F_r^{(1)}$ without any warping in between.

\subsection{Implementation}
During training, we perform self-reenactment and same identity reconstruction. For each video, we randomly sample a pair of source and driving frames and train the model with VGG-19 \cite{simonyan2014very} based perceptual loss \cite{johnson2016perceptual}, Patch-GAN \cite{isola2017image} and Hinge loss \cite{lim2017geometric} based adversarial loss, and discriminator feature matching loss between the real images and the synthesized images. We use $d=16$ as the latent code dimension and 5 pyramid levels for feature warping and re-alignment. We enable spectral normalization \cite{miyato2018spectral} for all convolutional layers in the model (including discriminator). We optimize the model with ADAM for $270k$ iterations. The training starts with an initial learning rate of $2 \times 10^{-4}$ and decreases to $2 \times 10^{-5}$ at $80k$ iterations and to $2 \times 10^{-6}$ at $160k$ iterations. Implementation details of network architecture and training loss functions can be found in the supplementary material.

\section{Experiments}
\subsection{Datasets and Baselines}
We use the VoxCeleb1 dataset \cite{nagrani2017voxceleb} for training, which contains more than 20k talking-head videos of over 1000 celebrities. We use the original train/test split provided by the authors to train and evaluate the proposed model. As a pre-processing step, we follow \cite{siarohin2019first,ren2021pirenderer} to crop the face and resize it to 256x256. We also fit the FLAME face model offline to all videos using DECA \cite{feng2021learning}. We also evaluate the VoxCeleb1 trained model on the VoxCeleb2 dataset \cite{chung2018voxceleb2}, which contains 5 times more identities than VoxCeleb1. We go through the same steps for data pre-processing and use the test split of VoxCeleb2 for evaluation.

The proposed model has two main variants depending on the geometric pattern in use: geometric displacement field or posed neural codes, denoted by \textit{``PWM + geom. disp.''} and \textit{``PWM + neural codes''} respectively. We compare both model variants with the state-of-the-arts including HeadGAN \cite{doukas2021headgan}, PIRender \cite{ren2021pirenderer}, FOMM \cite{siarohin2019first}, 3D-FOMM \cite{wang2021one}, Bi-Layer \cite{zakharov2020fast}, LSR \cite{meshry2021learned} and X2Face \cite{wiles2018x2face}. For all models except HeadGAN and 3D-FOMM, we adopt their officially released models for evaluation. Since no pre-trained model is available for HeadGAN, we follow its implementation details in \cite{doukas2021headgan} and train the model with FLAME as the underlying 3DMM. For 3D-FOMM, we use an unofficial implementation from \cite{3dfomm}. However, as noted by \cite{3dfomm}, they improve the original 3D-FOMM \cite{wang2021one} by adding SPADE blocks \cite{park2019semantic} in the decoder. Some baselines \cite{ren2021pirenderer, wiles2018x2face, zakharov2020fast, meshry2021learned} are originally designed for few-shot scenarios. To match our test configuration, we tested their models in one-shot setting using a single frame without finetuning on test subjects. All models except Bi-Layer \cite{zakharov2020fast} are trained on VoxCeleb1, which is trained on the larger VoxCeleb2 dataset. We benchmark all models on both VoxCeleb1 and VoxCeleb2.

\newcommand{\first}[1]{\textcolor{blue}{#1}}
\newcommand{\second}[1]{\textcolor{red}{#1}}
\setlength{\tabcolsep}{0.009\textwidth} 
\begin{table}[ht]
    \scriptsize
    \caption{Quantitative results. Best performing metric is shown in \first{blue}}
    \label{tab:results}
    \begin{tabular}{l|c|c|c|c|c|c|c|c|c}
    \toprule
    \multicolumn{10}{c}{VoxCeleb1} \\
    \hline
    \multicolumn{1}{c|}{} & \multicolumn{5}{c|}{Same Identity Reconstruction} & \multicolumn{4}{c}{Cross Identity Reconstruction} \\
    \hline
      & FID $\downarrow$ & CSIM $\uparrow$ & AKD $\downarrow$ &  AED $\downarrow$  & APD $\downarrow$ & FID $\downarrow$ & CSIM $\uparrow$ &  AED $\downarrow$  & APD $\downarrow$ \\
    \hline
    X2Face & 36.02 & 0.532 & 10.94 & 0.201 & 5.189 & 51.14 & 0.437 & 0.297 & 7.421 \\
    \hline
    Bi-Layer & 79.05 & 0.580 & 3.11 & 0.129 & 0.866 & 87.54 & 0.461 & 0.230 & 1.493 \\
    \hline
    LSR &  20.53 & 0.311 & 2.73 & 0.153  & 1.012 & 28.04 & 0.229 & 0.217 & 1.606 \\
    \hline
    FOMM & 10.87 & 0.788 & 2.25 & 0.087 & 0.688 & 30.84 & 0.562 & 0.218 & 1.612 \\
    \hline
    3D-FOMM & 5.47 & 0.792 & 2.23 & 0.091 & 0.770 & 22.34 & 0.605 & 0.247 & 2.149 \\
    \hline
    HeadGAN & 10.72 & 0.779 & 3.80 & 0.087 & 1.097 & 25.79 & 0.505 & 0.234 & 1.849 \\ 
    \hline
    PIRender & 9.47 & 0.745 & 3.41 & 0.119 & 1.148 & 23.82 & 0.530 & 0.222 & 1.916 \\
    \hline
    PWM + NMFC & 5.57 & 0.784 & 2.32 & 0.091 & 0.786 & 15.73 & 0.647 & 0.225 & 1.641 \\
    \hline
    PWM + geom. disp. & 3.95 & \first{0.794} & 2.29 & 0.081 & 0.756 & \first{14.19} & \first{0.653} &  0.228 & 1.668 \\
    \hline
    PWM + neural codes & \first{3.93} & 0.793 & 2.15 & 0.067 & 0.670 & 14.32 & 0.584 & 0.174 & \first{1.214} \\
    \hline
    PWM + geom. disp. + neural codes & 4.33 & 0.794 & \first{2.14} & \first{0.066} & \first{0.664} & 15.39 & 0.590 & \first{0.173} & 1.224 \\ 
    \hline
    PWM + geom. disp. + neural codes (BFM) & 4.80 & 0.779 & 4.04 & 0.087 & 1.110 & 16.78 & 0.546 & 0.233 & 1.956 \\
    \hline 
    \end{tabular}
    
    \begin{tabular}{l|c|c|c|c|c|c|c|c|c}
    \multicolumn{10}{c}{VoxCeleb2} \\
    \hline
    \multicolumn{1}{c|}{} & \multicolumn{5}{c|}{Same Identity Reconstruction} & \multicolumn{4}{c}{Cross Identity Reconstruction} \\
    \hline
      & FID $\downarrow$ & CSIM $\uparrow$ & AKD $\downarrow$ &  AED $\downarrow$  & APD $\downarrow$ & FID $\downarrow$ & CSIM $\uparrow$ &  AED $\downarrow$  & APD $\downarrow$ \\
    \hline
    X2Face & 30.34 & 0.538 & 21.66 & 0.188 & 7.780 & 41.31 & 0.379 & 0.280 & 9.757 \\
    \hline
    Bi-Layer & 57.26 & 0.541 & 2.84 & 0.138 & 1.153 & 66.22 & 0.436 & 0.221 & 1.780 \\
    \hline 
    LSR &  12.43 & 0.275 & 2.66 & 0.156 & 1.388 &  18.17 & 0.217 & 0.214 & 1.928 \\
    \hline
    FOMM & 8.32 & 0.718 & 2.69 & 0.118 & 1.291 & 26.00 & 0.475 & 0.231 & 2.545 \\
    \hline
    3D-FOMM & 3.49 & 0.723 & 3.11 & 0.118 & 1.498 & 14.52 & 0.567 & 0.240 & 3.189 \\
    \hline
    HeadGAN & 8.59 & 0.704 & 3.87 & 0.112 & 1.665 & 19.47 & 0.473 & 0.227& 2.647 \\
    \hline
    PIRender & 7.77 & 0.677 & 3.47 & 0.144 & 1.758 & 19.15 & 0.508 & 0.229 & 2.852 \\
    \hline
    PWM + NMFC & 4.17 & 0.725 & 2.64 & 0.118 & 1.162 & 10.68 & 0.601 & 0.230 & 2.142 \\
    \hline
    PWM + geom. disp. & \first{3.14} & \first{0.732} & 2.69 & 0.111 & 1.203 & \first{9.83} & \first{0.612} & 0.234 & 2.564 \\
    \hline
    PWM + neural codes & 3.29 & 0.723 & 2.37 & \first{0.087} & \first{0.954} & 10.15 & 0.528 & \first{0.177} & \first{1.498} \\
    \hline
    PWM + geom. disp. + neural codes & 3.18 & 0.726 & \first{2.36} & 0.088 & \first{0.954} & 9.97 & 0.541 & 0.178 & 1.533 \\
    \hline
    PWM + geom. disp. + neural codes (BFM) & 3.63 & 0.721  & 3.70  & 0.109  & 1.552 & 11.15 & 0.516 & 0.227 & 2.569 \\
    \bottomrule
    \end{tabular}
\vspace{-0.05\hsize}
\end{table}

\subsection{Evaluation Metrics}
We adopt \textit{Frechet Inception Distance} (FID) \cite{heusel2017gans}, \textit{Cosine Similarity} (CSIM) \cite{deng2019arcface}, \textit{Average Keypoint Distance} (AKD),  \textit{Average Pose Distance} (APD) and \textit{Average Expression Distance} (AED) as our evaluation metrics. FID measures the realism of the generated images and CSIM measures the capability of identity preservation. AKD, APD and AED evaluate the geometry accuracy in terms of facial landmarks, head pose (yaw, pitch \& roll) and the expression parameters obtained from FLAME. More details of evaluation metrics can be found in the supplementary material.

For same identity reconstruction, we perform self-reenactment where the source portrait image and the driving portrait image come from the same video clip. During evaluation, we randomly sample pairs of source and driving frames from each video and do both forward and backward reconstruction. This is to ensure that we always cover the difficult cases. For instance, synthesizing a frontal face from a profile face is always harder than the reversed due to severe occlusions in profile face. 

For cross identity reconstruction, we perform face reenactment where the source portrait image and the driving portrait image come from different videos. One advantage of our model is that the facial motion transfer can be accomplished independent of identity due to the disentangled parameterization of shape, pose and expression in FLAME. Therefore, we replace the shape of the driving face with that of the source to preserve the facial geometry. Additionally, because facial landmarks reflect person-specific facial geometry and the driving image has a different identity and appearance from the source image, AKD can no longer be used to measure the accuracy of motion transfer on this task. 

\begin{figure}[h]
\vspace{-0.02\hsize}
\scriptsize
  \centering
  \includegraphics[width=\textwidth]{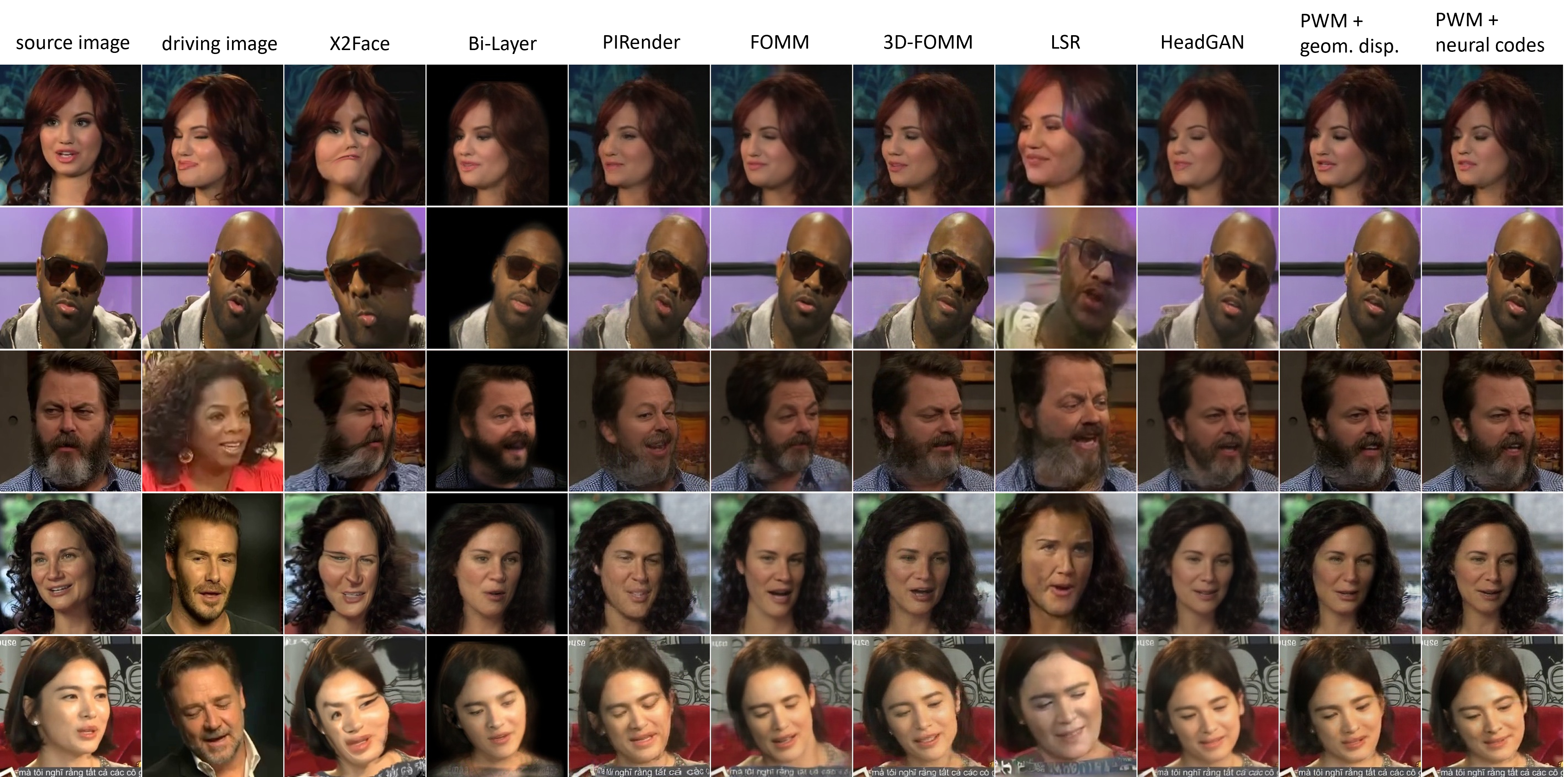}
  \caption{Qualitative results. Top two rows are same identity reconstruction and the bottom three rows are cross identity reconstruction. Zoom in for better visual comparison.}
  \label{fig:rec_imgs}
\vspace{-0.05\hsize}
\end{figure}

\subsection{Results}
Qualitative results are presented in Fig. \ref{fig:rec_imgs}. We can see that our models generate images with better realism and higher sharpness compared to other baselines. We refer readers to the supplementary material for more qualitative results. Quantitative results are shown in Tab. \ref{tab:results}. Our method outperforms prior models on all metrics. X2Face and Bi-Layer have relatively worse numbers on all metrics. This is because they encode motion information in an embedding vector, losing spatial information that could otherwise be useful to guide their model to generate quality images. FOMM has good results on same identity reconstruction, but its performance drops significantly on cross identity reconstruction. This might be FOMM's motion descriptor dependent on the subject's appearance. Once the subject's identity changes, FOMM brings not only facial motion but also the information of the new identity to image synthesis. Compared to FOMM, 3DMM based models such as HeadGAN and PIRender show better results on this task due to 3DMM's disentangled representation of identity and facial movement. 3D-FOMM outperforms other baselines significantly, which also justifies the inclusion of 3D information for guidance. Our models achieve the best results attributed to the proposed PWM and guiding geometric patterns. We also notice that \textit{``PWM + geom. disp.''} is better on image quality (FID \& CSIM) while \textit{``PWM + neural codes''} performs better on geometry accuracy (AKD, AED \& APD), suggesting their slightly different focus in guidance, which we will analyze individually in Sec. \ref{sec:ablation}. 

\vspace{-0.02\hsize}
\subsection{Face Editing}

\begin{wrapfigure}{r}{0.6\linewidth}
\scriptsize
\includegraphics[width=\linewidth]{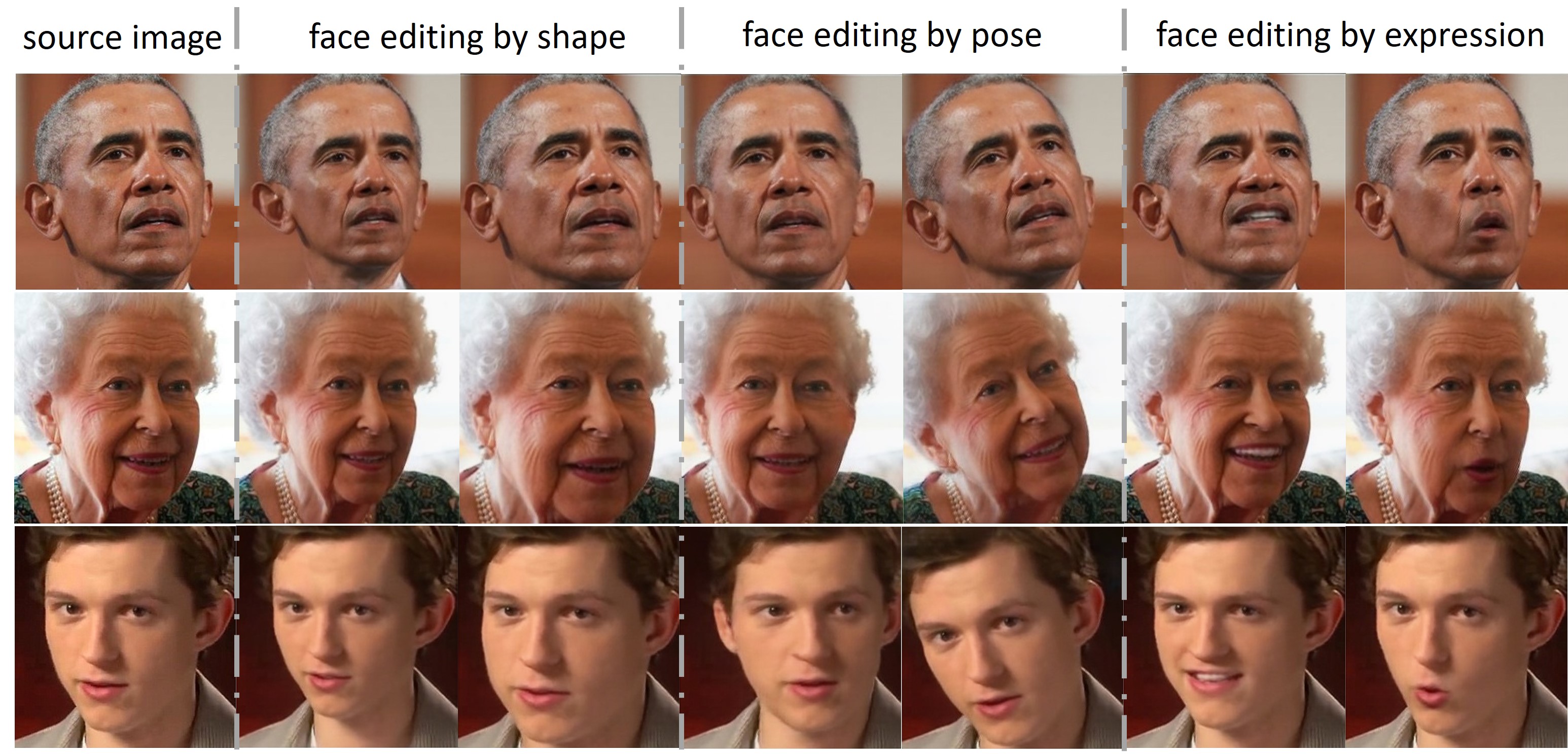}
\caption{Face editing results.}
\label{fig:face_edit}
\vspace{-0.04\hsize}
\end{wrapfigure} 

The disentangled parameterization of FLAME enables us to edit the shape, pose and expression individually to generate customized portrait images. Fig. \ref{fig:face_edit} shows some examples of face editing. Our model generates photo-realistic images with natural poses and expressions. It can even synthesize teeth in the open mouth region (see  second to last column of Fig. \ref{fig:face_edit}).

\subsection{Ablation Study} \label{sec:ablation}
Two major components that have a significant impact on the model performance are the PWM and the guiding geometric patterns. Below we assess their individual influence. 

\textbf{The effectiveness of PWM}. We train a model variant similar to \textit{``PWM + neural codes''} but replace the posed neural codes with the normalized mean face coordinates (NMFC). We render a $3$-channel image using the normalized coordinates of the mean face obtained from FLAME. The same representation has been used in HeadGAN \cite{doukas2021headgan}. We denote this model by \textit{``PWM + NMFC''}. We compare this model variant with HeadGAN, which solely predicts the displacement field at a single pyramid scale and uses that to warp features. As shown in Tab. \ref{tab:results}, \textit{``PWM + NMFC''} outperforms HeadGAN by a large margin on all metrics except AED, especially on the image quality metrics FID and CSIM. This shows the importance of feature realignment at each pyramid level and the effectiveness of alternating feature warping and displacement prediction via a sequence of PWMs.

\textbf{The role of geometric patterns}. By comparing \textit{``PWM + geom. disp.''} and \textit{``PWM + neural codes''} in Tab. \ref{tab:results}, we find that the geometric displacement field tends to generate images with higher fidelity and is better at preserving identity, but at the sacrifice of geometry accuracy. There are several reasons for this. First, unlike posed neural codes, geometric displacement field does not encode facial geometry and semantics. Instead, it only provides information about \textbf{relative} motion transfer, which makes it harder to generate images spatially aligned with the target pose, hence lower geometry accuracy. With the same architecture and complexity, weaker constraint on facial geometry and semantics makes \textit{``PWM + geom. disp.''} focus more on realism during training, resulting in better image quality. Second, a model needs to synthesize more pixels for more aggressive motion change. For example, when the model is used for face frontalization, the more it tries to correct the face pose, the more pixels it needs to create for the originally occluded face region. By doing less motion change (less pixel synthesis), \textit{``PWM + geom. disp.''} naturally obtains better image quality. This explains the trade-off between geometry correctness and image quality.


The above can also be justified by metrics such as L1, SSIM and LPIPS. Because those metrics are calculated pixelwise and averaged by spatial dimensions, they, to certain extent, reflect how accurate the synthesized images are spatially aligned with the ground truth. We compute the metrics for same identity reconstruction where ground truth image of the source identity is available. Results are presented in Tab. \ref{tab:ablation}. The better results of \textit{``PWM + neural codes''} can be attributed to the posed neural codes providing more information about facial geometry and semantics at the corresponding pixel locations. 

\setlength{\tabcolsep}{0.015\textwidth} 
\begin{wraptable}{r}{0.5\linewidth}
    \scriptsize
    \caption{Additionally metrics for same identity reconstruction.}
    \label{tab:ablation}
    \centering
    
    \begin{tabular}{l|c|c|c}
    \toprule
    \multicolumn{4}{c}{VoxCeleb1} \\
    \hline
      & L1 $\downarrow$ & SSIM $\uparrow$ & LPIPS $\downarrow$ \\
    \hline
    PWM+geom. disp. & 11.93 & 0.779 & 0.106 \\
    \hline
    PWM+neural codes & \first{11.38} & \first{0.782} & \first{0.103} \\
    \hline
    \end{tabular}
    
    \begin{tabular}{l|c|c|c}
    \multicolumn{4}{c}{VoxCeleb2} \\
    \hline
      & L1 $\downarrow$ & SSIM $\uparrow$ & LPIPS $\downarrow$ \\
    \hline
    PWM+geom. disp. & 13.60 & 0.743 & 0.137 \\
    \hline
    PWM+neural codes & \first{13.04} & \first{0.747} & \first{0.133} \\
    \bottomrule
    \end{tabular}
\vspace{-0.02\hsize}
\end{wraptable}

We also try the combination of geometric displacements and neural codes (via channel concatenation) as guidance, shown as \textit{``PWM + geom. disp. + neural codes''} in Tab. \ref{tab:results}. Compared to models that use geometric displacements and neural codes separately, combining the two does not boost the performance further. Like neural codes guided model,  the mixed model tends to favor geometric correctness by trading off realism. 

\textbf{The dependence on 3DMM}. We are interested in the dependence of our method on the type of 3DMM in use. Therefore, we add an experiment with the ``PWM + geom. disp. + neural codes'' configuration but use the Bazel Face Model (BFM) \cite{blanz1999morphable, deng2019accurate, paysan20093d} as the underlying 3DMM. As can been seen in Tab. \ref{tab:results}, BFM underperforms FLAME but it still outperforms other baselines on most metrics, demonstrating that our method is generic. We also extend AED and APD by fitting a BFM face to compute the expression and pose parameters. Results are shown in Tab. \ref{tab:BFM}. The FLAME based model has better performance even on the BFM computed AED and APD. Compared to BFM that only models the face region, we believe the better quality of FLAME benefits from its modeling of full head and neck. 

\setlength{\tabcolsep}{0.013\textwidth} 
\begin{table}[ht]
    \scriptsize
    \caption{Comparison between different 3DMMs. The subscripts $F$ and $B$ denote a metric computed using FLAME and BFM respectively.}
    \label{tab:BFM}
    \centering
    \begin{tabular}{l|c|c|c|c|c|c|c|c}
    \toprule
    \multicolumn{9}{c}{VoxCeleb1} \\
    \hline
    \multicolumn{1}{c|}{} & \multicolumn{4}{c|}{Same Identity Reconstruction} & \multicolumn{4}{c}{Cross Identity Reconstruction} \\
    \hline
    & AED$_F$  & APD$_F$ & AED$_B$  & APD$_B$ & AED$_F$  & APD$_F$ & AED$_B$  & APD$_B$ \\
    \hline
    PWM + geom. disp. + neural codes (FLAME) & \first{0.066} & \first{0.664} & 0.088 & \first{0.791} & \first{0.173} & \first{1.224} & \first{0.228} & \first{1.702} \\
    \hline
    PWM + geom. disp. + neural codes (BFM) & 0.087 & 1.110 & \first{0.087} & 1.245 & 0.233 & 1.956 & 0.230 & 2.128 \\
    \hline 
    \end{tabular}
    
    \begin{tabular}{l|c|c|c|c|c|c|c|c}
    \multicolumn{9}{c}{VoxCeleb2} \\
    \hline
    \multicolumn{1}{c|}{} & \multicolumn{4}{c|}{Same Identity Reconstruction} & \multicolumn{4}{c}{Cross Identity Reconstruction} \\
    \hline
    & AED$_F$  & APD$_F$ & AED$_B$  & APD$_B$ & AED$_F$  & APD$_F$ & AED$_B$  & APD$_B$  \\
    \hline
    PWM + geom. disp. + neural codes (FLAME) & \first{0.088} & \first{0.954} & 0.116 & \first{1.366} & \first{0.177} & \first{1.533} & 0.241 & \first{2.294} \\
    \hline
    PWM + geom. disp. + neural codes (BFM) & 0.109 & 1.552 & \first{0.110} & 1.937 & 0.227 & 2.569 & \first{0.238} & 2.885 \\ 
    \bottomrule
    \end{tabular}
\end{table}

\textbf{Further discussion}. One limitation is the dependence on the fitting quality of 3DMM. In fact, this is a common problem because almost all models rely on either 3DMM fitting or landmark detection. Isolating and quantifying the impact of 3DMM fitting and landmark detection still remains an open question. However, we have seen \cite{siarohin2019first,wang2021one} made efforts in using self-learned landmarks (which the model deems important but are less interpretable to human) for face warping. Similarly, one future direction could be embedding a self-hosted 3DMM within the model to remove the dependence on external 3DMM fitting.

\vspace{-0.03\hsize}
\section{Conclusion}
A novel geometry driven model is presented for one-shot face animation. We show that our model outperforms other baselines in both image quality and geometry accuracy. By studying each proposed component, we demonstrate the effectiveness of the progressive warping module (PWM) and find out that each geometric pattern has a different focus in guidance. The geometric displacement field achieves better image quality whereas the posed neural codes favor better geometry correctness.

\clearpage
\bibliography{egbib}
\end{document}


\maketitle

\section{Training Losses}
In this section we provide the details of the loss functions that are used during training. We denote the ground truth driving image by $I_d$ and generated output image by $I_o$.

\subsection{Perceptual loss} 
We adopt a VGG-19 \cite{simonyan2014very} based perceptual loss \cite{johnson2016perceptual}. A pyramid of three scales (256x256, 128x128, 64x64) of input images are used and the loss is calculated on the convolutional outputs (instead of the pooling outputs) of the first five blocks. 

\begin{equation}
\mathcal{L_P}(I_d, I_o)=\sum_{s}\sum_{j}\sum_{i}|\mathcal{V}_{j, i}(I_d^{s}) - \mathcal{V}_{j, i}(I_o^{s})|,
\end{equation}

where $s$ denotes the image scale and $\mathcal{V}_{j, i}(\cdot)$ is the $j$-th layer output of VGG-19 at spatial location $i$.

\subsection{Adversarial loss} 
To generate photo realistic images, we train the model with Patch-GAN \cite{isola2017image} and Hinge loss \cite{lim2017geometric}. Similar to perceptual loss, we apply adversarial training on a pyramid of three scales (256x256, 128x128, 64x64) of input images.

\begin{eqnarray}
\mathcal{L_G} &=& -\sum_s \sum_i \mathcal{D}_i(I_o^{s}), \\
\mathcal{L_D} &=& -\sum_s \sum_i \left [ min(0, -1-\mathcal{D}_i(I_o^{s})) + min(0, -1+\mathcal{D}_i(I_d^{s})) \right ],
\end{eqnarray}

where $s$ denotes the image scale, $i$ is the spatial location of Patch-GAN discriminator output, $\mathcal{L_G}$ and $\mathcal{L_D}$ are the generator loss and discriminator loss respectively.

\subsection{Feature matching loss}
To stabilize adversarial training, we also adopt discriminator feature matching loss between the real images and generated images. Again, the loss is calculated on a pyramid of three image scales.

\begin{equation}
\mathcal{L_M}(I_d, I_o)=\sum_{s} \sum_{j}\sum_{i}|\mathcal{D}_{j, i}(I_d^{s}) - \mathcal{D}_{j, i}(I_o^{s})|,
\end{equation}

where $s$ denotes the image scale, $\mathcal{D}_{j, i}(\cdot)$ denotes the $j$-th layer output of the discriminator $\mathcal{D}$ at spatial location $i$.

\subsection{Additional warping constraints}
So far we have described how we compute the losses for the generated output image $I_o$. To add more constraints for the model to learn correct displacements, we take the predicted displacement field from the last pyramid in the decoder and use that to warp the source image to create a warped image $I_w$. This image is less realistic than $I_o$, but still can provide some additional supervision to train the model. Therefore, we take the warped image $I_w$ and compute all of the above losses between $I_w$ and $I_d$. 

\section{Implementation Details}
\begin{figure}[h]
\scriptsize
  \centering
  \includegraphics[width=\textwidth]{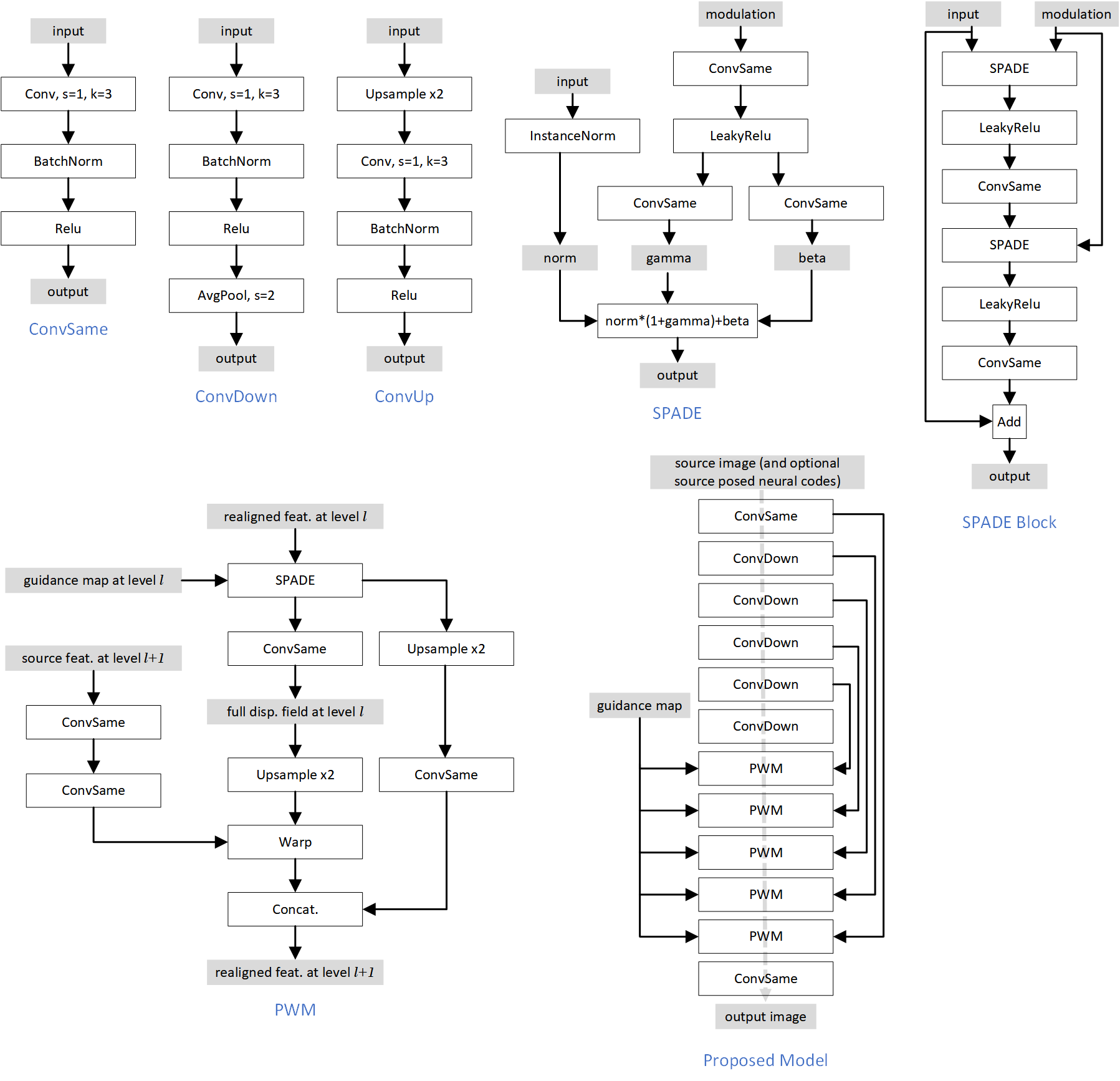}
  \caption{Implementation details of the proposed model.}
  \label{fig:net_diag}
\end{figure}

Here we show the implementation details of the proposed model in Fig. \ref{fig:net_diag}. The model adopts an encoder-decoder structure. For the model variant \textit{``PWM + geom. disp.''}, the guidance map shown in the figure would be the rendered geometric displacement field and the input would be the source image only. For the model variant \textit{``PWM + neural codes''}, the posed neural codes (rendered in target image space) would be used as the guidance map. In this case, we also render the latent codes in source image space and concatenate it with the source image before being fed to the encoder. 

\section{More on Evaluation Metrics}
We use the below metrics to evaluate our model and other baselines.

\textit{Frechet Inception Distance} (FID) \cite{heusel2017gans}. We adopt InceptionV3 \cite{szegedy2016rethinking} as the backbone network and use the output from the last average pooling layer as the input feature vectors to FID. FID measures how close the distribution of generated images is to that of the real images. Therefore it reflects the realism of generated images. For same identity reconstruction, we compute FID between generated images and ground truth target images. For cross identity reconstruction, we compute FID between generated images and input source images. This is to ensure that we measure the realism based on the same identity.

\textit{Cosine Similarity} (CSIM). We adopt ArcFace \cite{deng2019arcface} as the underlying network to produce the face embedding for comparison. We first detect 68 facial landmarks with \cite{pytorchfa, bulat2017far} and align the face with a predefined reference face to address any ambiguity in scale and rotation. Then we calculate CSIM on the aligned face images. Because the underlying network is pre-trained for face verification, CSIM can be used to evaluate a model's capability of preserving identity. Similar to FID, we compute CSIM between generated images and ground truth target images for same identity reconstruction and compute CSIM between generated images and input source images for cross identity reconstruction.

\textit{Average Keypoint Distance} (AKD). We detect 68 facial landmarks \cite{pytorchfa, bulat2017far} and compute the L1 distance between the landmarks of generated images and ground truth target images. AKD is used as a way of measuring the accuracy of motion transfer. It is only used for same identity reconstruction because we do not have the reenacted image of the source identity available as ground truth. Computing AKD between generated images and driving images of different identities is erroneous since facial landmarks encode not only pose information but also person-specific shape information. We should exclude any noise from identities by using the same subject.

\textit{Average Pose Distance} (APD) and \textit{Average Expression Distance} (AED). We fit FLAME on images and compute the L1 distances of their pose and expression parameters respectively. Note that the pose here refers to the yaw, pitch and roll angles of the global head pose estimated by FLAME. APD and AED are used to evaluate how accurate the model is to transfer the head pose and facial expression. For same identity reconstruction, we compute APD and AED between generated images and the ground truth target images. For cross identity reconstruction, we compute APD and AED between generated images and driving images because they are expected to have the same poses and expressions.

\section{Additional Qualitative Results}
\begin{figure}
    \centering
    \includegraphics[width=\textwidth]{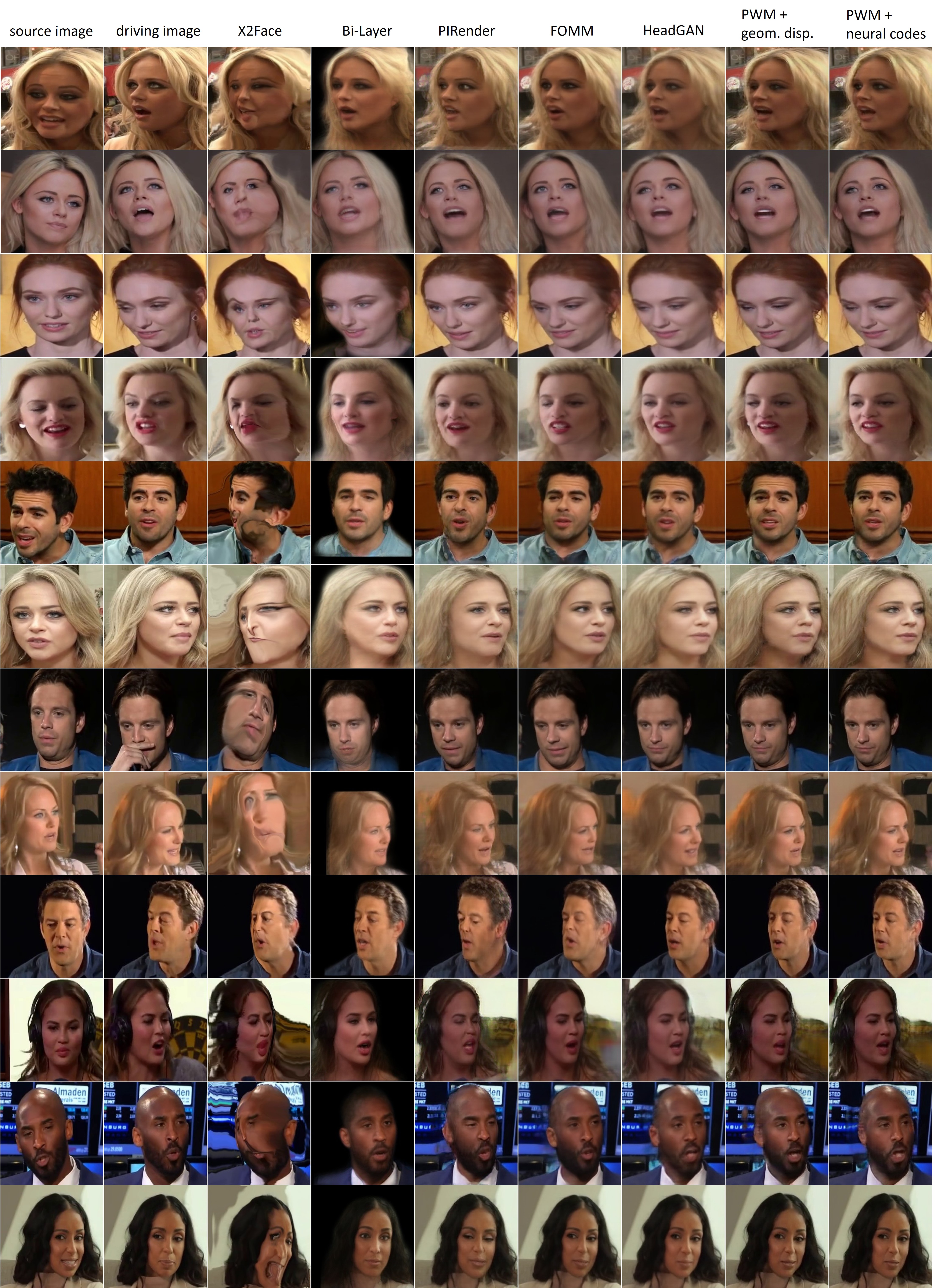}
    \caption{Qualitative results for same identity reconstruction.}
    \label{fig:same_id_rec_supp}
\end{figure}

\begin{figure}
    \centering
    \includegraphics[width=\textwidth]{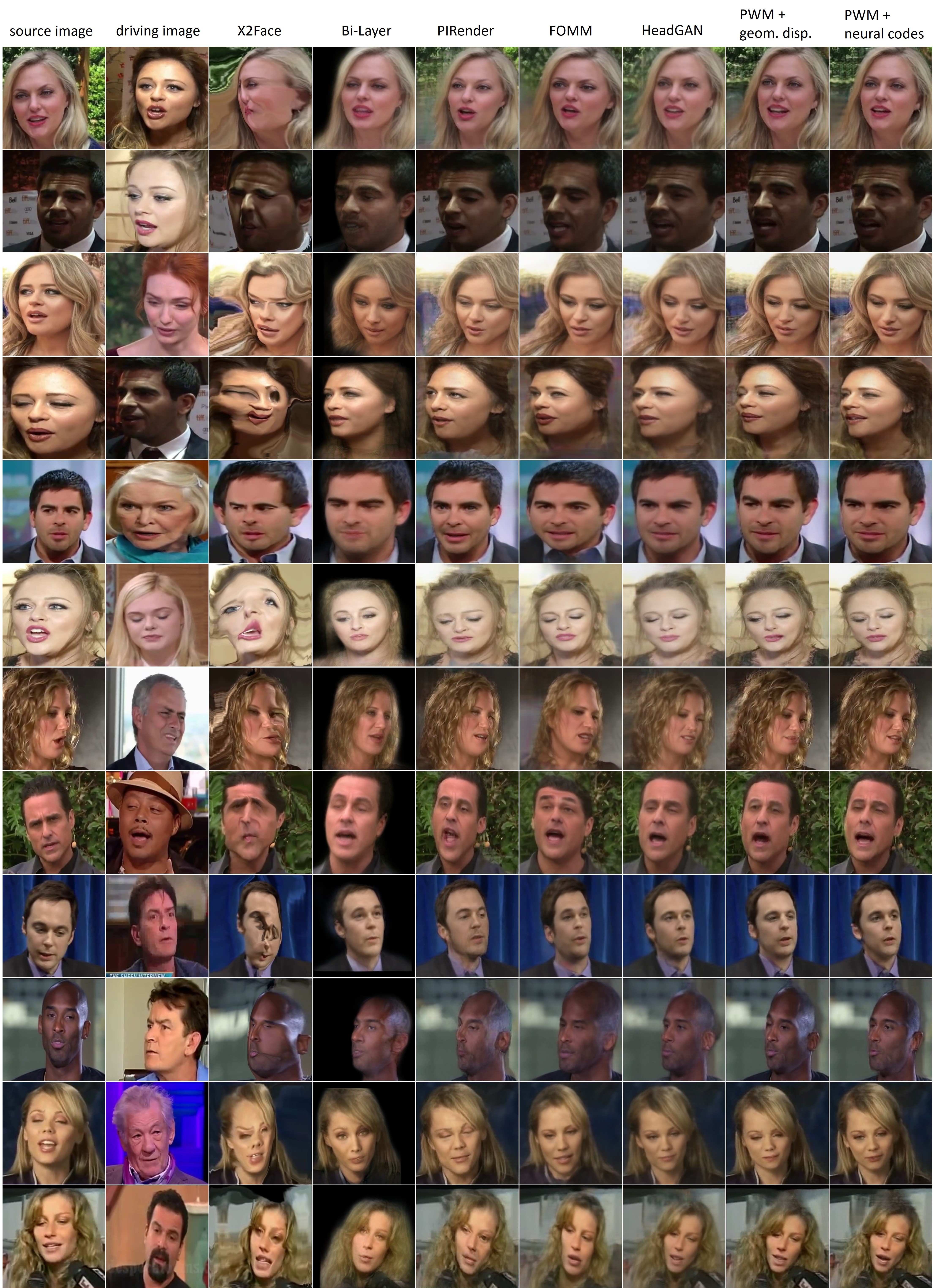}
    \caption{Qualitative results for cross identity reconstruction.}
    \label{fig:cross_id_rec_supp}
\end{figure}

In this section, we present more qualitative results to show the visual quality of synthesized portrait images. See Fig. \ref{fig:same_id_rec_supp} for same identity reconstruction and Fig. \ref{fig:cross_id_rec_supp} for cross identity reconstruction.

\clearpage
\bibliography{egbib}